\documentclass[letterpaper, 10 pt, conference]{ieeeconf}  

\IEEEoverridecommandlockouts

\usepackage{amsmath, amsfonts,bm}

\def\1{\bm{1}}

\DeclareMathAlphabet{\mathsfit}{\encodingdefault}{\sfdefault}{m}{sl}
\SetMathAlphabet{\mathsfit}{bold}{\encodingdefault}{\sfdefault}{bx}{n}

\usepackage{microtype}
\usepackage{amssymb}
\usepackage{graphicx}
\usepackage{stfloats}
\usepackage[ruled,vlined,noend]{algorithm2e}
\usepackage{booktabs, multirow, multicol}
\usepackage{balance}
\usepackage[colorlinks]{hyperref}
\usepackage{color}
\usepackage{float}
\usepackage{gensymb}
\usepackage{caption}
\usepackage[usenames,dvipsnames]{xcolor}
\let\labelindent\relax
\usepackage{enumitem}
\usepackage{cuted} 
\newcommand{\bl}[1]{{\flushleft\textbf{#1}}}
\newcommand{\meanstd}[2]{$#1${\scriptsize$\;\pm\,#2$}}

\usepackage{pifont}

\title{Rapidly Learning Soft Robot Control \\ via Implicit Time-Stepping}

\author{Andrew Choi$^1$, Dezhong Tong$^2$, Xiaonan Huang$^{2,\dagger}$
\thanks{$^1$Andrew Choi is with Horizon Robotics, Cupertino, CA, USA. {\tt\small asjchoi@cs.ucla.edu}}
\thanks{$^2$Dezhong Tong and Xiaonan Huang are with the Robotics Department at the University of Michigan, Ann Arbor, MI, USA. {\tt\small dezhong@umich.edu, xiaonanh@umich.edu}}
\thanks{$^\dagger$Corresponding author.}
}

\begin{document}

\maketitle

\begin{figure*}[t]
\centering
\includegraphics[width=0.9\textwidth]{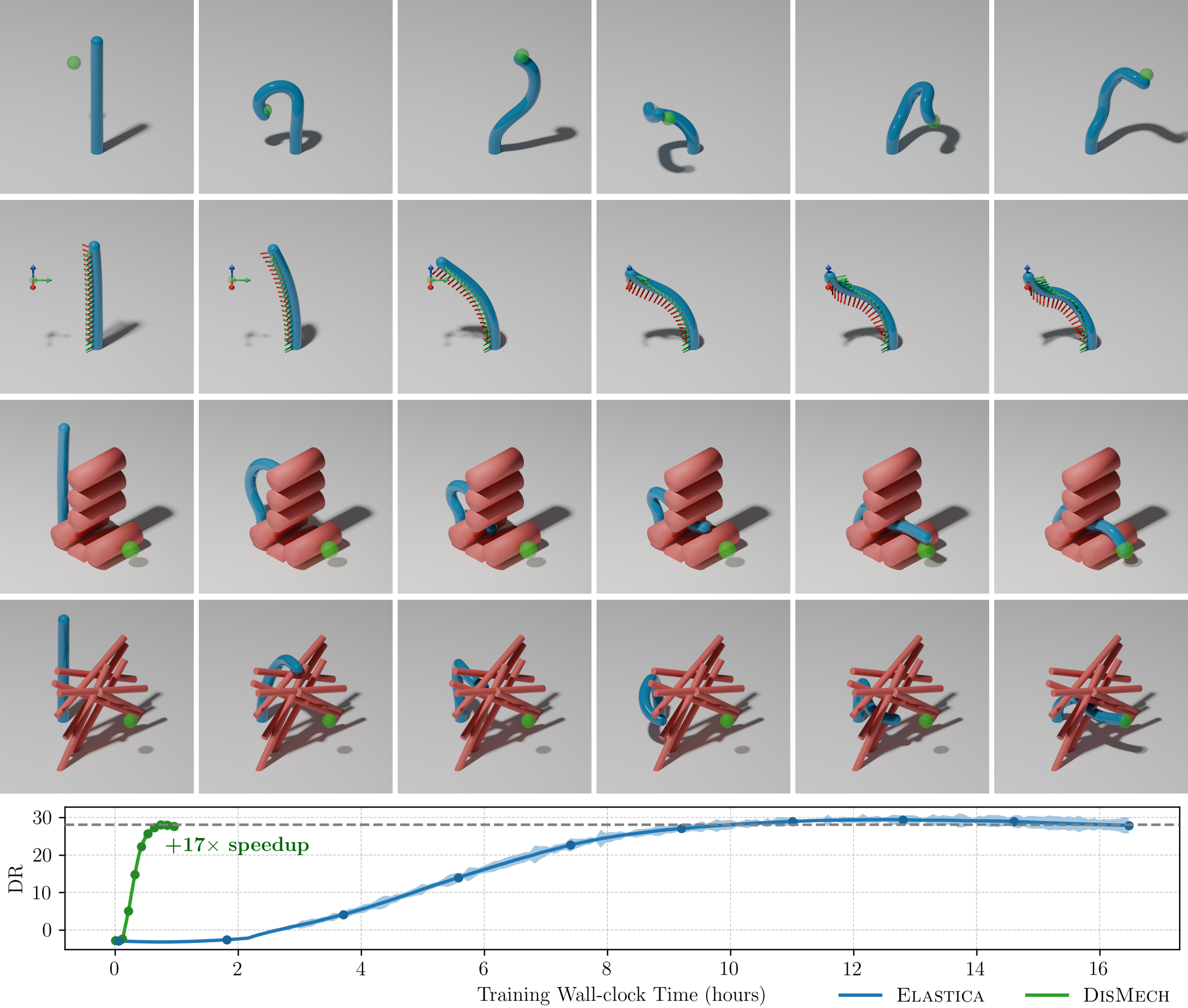}
\caption{
Visualization of four soft manipulator policies.
\textbf{Row~1:} end-effector tracking of a moving target traveling at 0.5\,m/s.
\textbf{Row~2:} 4D inverse kinematics for position and yaw.
\textbf{Row~3:} 2D static target reaching through a set of tight obstacles.
\textbf{Row~4:} static target reaching through eight random 3D obstacles.
Note how the policy first attempts to take the shortest path to the target sphere and upon encountering resistance, starts to probe for gaps within the obstacles, demonstrating emergent tactile behavior.
\textbf{Bottom:} average discounted return (DR) versus wall-clock time for the 3D contact case using \textsc{Elastica} and \textsc{DisMech} as the simulator. \textsc{DisMech} attains over $17\times$ faster training per iteration.
}
\label{fig:snapshot}
\end{figure*}

\begin{abstract}
With the explosive growth of rigid-body simulators, policy learning in simulation has become the de facto standard for most rigid morphologies.
In contrast, soft robotic simulation frameworks remain scarce and are seldom adopted by the soft robotics community.
This gap stems partly from the lack of easy-to-use, general-purpose frameworks and partly from the high computational cost of accurately simulating continuum mechanics, which often renders policy learning infeasible.
In this work, we demonstrate that rapid soft robot policy learning is indeed achievable via implicit time-stepping.
We use \textsc{DisMech}, which is a general-purpose, fully implicit soft-body simulator capable of handling both soft dynamics and frictional contact.
We further introduce delta natural curvature control, a method analogous to delta joint position control in rigid manipulators, providing an intuitive and effective means of enacting control for soft robot learning.
To highlight the benefits of implicit time-stepping and delta curvature control, we conduct extensive comparisons across four diverse soft manipulator tasks against one of the most widely used soft-body frameworks, \textsc{Elastica}.
With implicit time-stepping, parallel stepping of 500 environments achieves up to $6\times$ faster speeds for non-contact cases and up to $40\times$ faster for contact-rich scenarios.
Finally, cross-simulator sim-to-sim evaluation shows that these gains are achieved while maintaining comparable task-level performance, suggesting that implicit time-stepping can significantly improve the practicality of policy learning for soft robots.
\end{abstract}

\section{Introduction}
Rigid-body simulators have become a central tool for robot policy learning, with strong results demonstrated across platforms such as MuJoCo~\cite{emanuel_2012_mujoco}, IsaacSim~\cite{makoviychuk2021isaacgymhighperformance}, and Genesis~\cite{Genesis}.
In contrast, simulator-driven policy learning is far less common in soft robotics, where many studies remain focused on hardware demonstrations, model-based control, or task-specific systems.
Two factors contribute to this gap: the limited availability of easy-to-use, general-purpose soft-body simulators, and the high computational cost of accurately simulating continuum mechanics and frictional contact.
As a result, it remains unclear whether soft-body simulators can support policy learning at the scale and speed now common in rigid-body robotics.

Unlike rigid manipulators, which often have only a small number of generalized coordinates, soft robots are continuum systems whose dynamics must be approximated through spatial discretization. Capturing elasticity, large deformation, and contact typically requires many degrees of freedom and can impose severe timestep restrictions in explicit simulation. These factors make soft-robot policy learning challenging both computationally and algorithmically, because expensive environment stepping directly limits training throughput.

In this work, we show that rapid policy learning for soft manipulators is possible using implicit time-stepping.
Our approach combines two ingredients: (1) aggressive implicit integration using the fully implicit simulator \textsc{DisMech}~\cite{choi_dismech}, and (2) delta natural curvature control, a soft-body analogue of delta joint control that provides an intuitive and learning-friendly action parameterization. While prior work on \textsc{DisMech} demonstrated speed advantages over explicit methods on canonical mechanics problems, our focus here is on whether these simulation advantages translate into faster policy learning.

To answer this question, we evaluate four representative soft-manipulator tasks and compare against \textsc{Elastica}~\cite{gazzola2018forward}, a widely used soft-body simulation framework based on a related reduced-order centerline representation. Using matched task setups and shared state/action interfaces with minor modifications, we find that \textsc{DisMech} preserves comparable task-level behavior while substantially reducing simulation and training cost. Our results suggest that implicit time-stepping can significantly improve the practicality of soft-robot policy learning, while also providing a useful benchmark for evaluating future soft-body simulators in learning settings.

The contributions of this work are as follows:
\begin{enumerate}[leftmargin=12pt, topsep=0pt, itemsep=0pt]
    \item We present a systematic comparison of policy learning performance in \textsc{DisMech} and \textsc{Elastica} across four representative soft manipulator tasks. Because both simulators use related reduced-order centerline representations, they provide a natural basis for controlled comparison under closely matched state and action formulations.
    
    \item We introduce \textit{delta natural curvature control}, a soft-body analogue of delta joint control in rigid manipulators, and demonstrate that it provides an intuitive and effective action parameterization for learning soft robot control policies.
    
    \item We show that implicit time-stepping can substantially accelerate soft robot policy learning, particularly in contact-rich settings, while maintaining comparable task-level behavior in cross-simulator sim-to-sim evaluation.
    
    \item We provide an open-source benchmark and implementation for studying both learning algorithms and simulator effects in soft robot policy learning.\footnote{Code: \url{https://github.com/QuantuMope/dismech-rl}}
\end{enumerate}

\section{Related Work}
\label{sec:related_work}
Soft-robot simulation spans a range of representations that trade off physical fidelity, computational cost, and compatibility with control and learning pipelines.
For slender soft manipulators, a common abstraction is the deformable linear object (DLO), whose behavior can often be captured effectively by one-dimensional reduced-order models rather than full volumetric discretizations.
Early approaches approximated elasticity using rigid-body chains connected by compliant spring--damper elements~\cite{graule2020somo}, but such models are limited in representing coupled deformation modes such as twist, shear, and buckling~\cite{tong_2023_snap_buckling, tong2025snapjumping}.
At the other end of the spectrum, the finite element method (FEM) provides a general framework for soft-body simulation, with platforms such as SOFA~\cite{Faure2012sofa} supporting soft-robot modeling and control~\cite{coevoet_2017_sofa_robot}; however, FEM-based methods often require high-resolution volumetric meshes and correspondingly high computational cost, motivating reduced-order and asynchronous techniques for real-time and learning-based applications~\cite{zhang2019modeling, largilliere_2015_asynch_fem}.

For slender structures, reduced-order rod models often provide a favorable balance between physical fidelity and computational efficiency.
Much of this literature builds on either the Kirchhoff rod model~\cite{kirchhoff1859uber} or the Cosserat rod model~\cite{cosserat}.
Discrete differential geometry (DDG)-based formulations of Kirchhoff rods, such as Discrete Elastic Rods (DER)~\cite{bergou2008der, bergou2010dvt}, provide an efficient and geometrically consistent framework for simulating slender deformable structures and soft robots~\cite{tong2025discrete, huang2025tutorial}.
These methods have supported a range of soft-robot locomotion, including jumping~\cite{tong2025snapjumping}, rolling~\cite{huang2020rolling}, and  swimming~\cite{huang2021star_swimming, hao2024tumbling_attitude}.
Building on this line of work, \textsc{DisMech}~\cite{choi_dismech} combines DER with implicit contact handling~\cite{choi_imc_2021, tong_imc_2022}, yielding a fully implicit time-stepping framework for stable simulation under large deformation and frictional contact.
A widely used alternative is \textsc{Elastica}~\cite{gazzola2018forward}, which adopts a Cosserat-rod formulation and has become a common simulator for soft-robot control and learning.
Other reduced-order frameworks include SoroSim~\cite{mathew2022sorosim}, while broader platforms such as Genesis~\cite{Genesis} reflect growing interest in unified rigid--soft simulation.

Reinforcement learning has increasingly been explored for soft-robot control in both model-free and model-based settings.
Early studies demonstrated RL-based control of soft manipulators and PneuNet-style systems~\cite{you2017model, zhang2017toward}, followed by deep RL for continuum-arm control~\cite{satheeshbabu2019open} and model-based RL for dynamic soft manipulators~\cite{thuruthel2018model}.
At the simulator level, \textsc{Elastica} has been coupled with standard RL algorithms for compliant-arm control~\cite{Naughton2021_elastica_rl}, while SofaGym~\cite{schegg2023sofagym} provides a broader learning platform built on SOFA.
More recent work has also demonstrated agile soft-arm maneuvers and sim-to-real transfer using simulation-trained RL~\cite{jitosho2023reinforcement, wei2023axis}.
Despite this progress, comparatively less attention has been given to how simulator design choices---and in particular time integration strategy---affect policy learning throughput and task-level performance.
This paper addresses this question by comparing the implicit simulator \textsc{DisMech} with the widely used framework \textsc{Elastica} across shared soft-manipulation tasks, while also evaluating a learning-oriented action parameterization based on delta natural curvature control.

\section{Methodology}
\label{sec:methodology}

\begin{figure}
\centering
\includegraphics[width=\columnwidth]{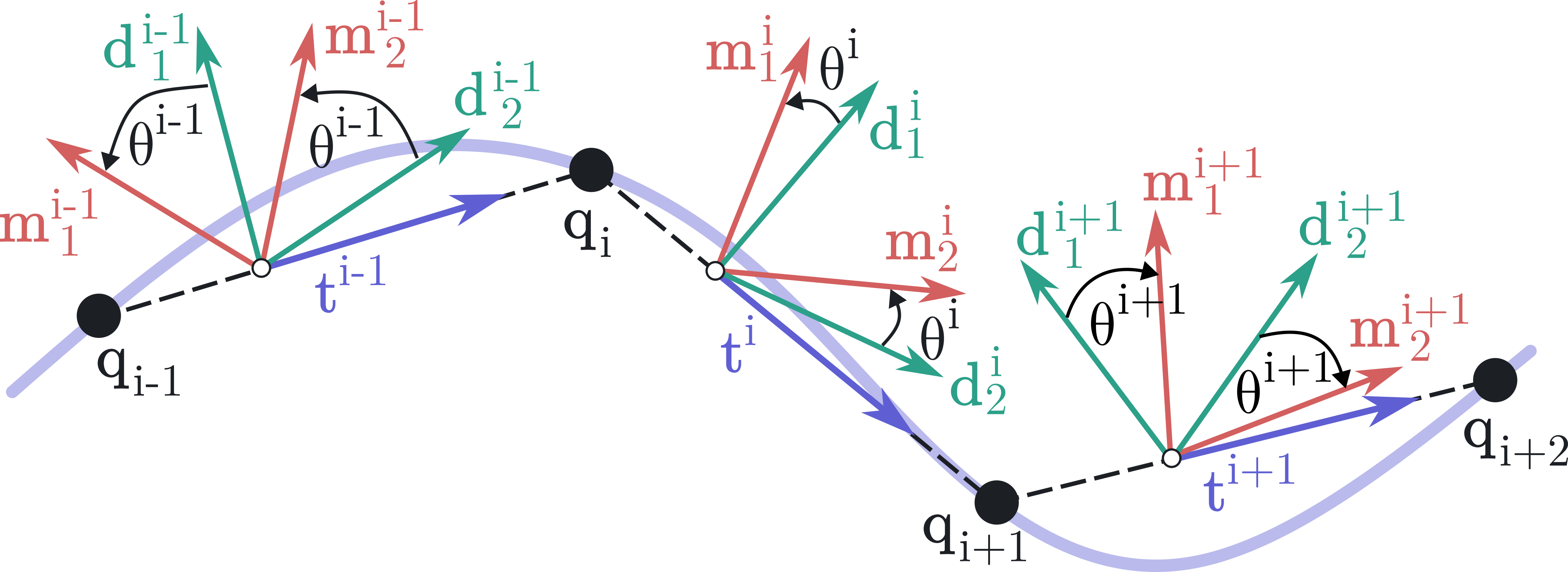}
\caption{Discrete rod schematic. A continuous centerline is discretized into nodes $\mathbf q_i$. Each discrete edge carries a reference frame $\{\mathbf d^i_1, \mathbf d^i_2, \mathbf t^i\}$ and a material frame $\{\mathbf m^i_1, \mathbf m^i_2, \mathbf t^i\}$, which are used to characterize bending and twisting deformations at interior nodes.}
\label{fig:centerline}
\end{figure}

In this section, we summarize the elastic energy formulation of \textsc{DisMech} and the control parameterization used in our learning framework.
For brevity, we omit the full equations of motion and external-force models, including contact, and refer the reader to the original \textsc{DisMech} paper~\cite{choi_dismech} for details.

\subsection{Discrete Centerline Representation}

We first describe the reduced-order representation of the discrete centerline.
An elastic rod’s discrete centerline consists of $N$ vertices, each represented by a Cartesian position $(\mathbf{q}_i)_{i=0}^{N-1}$.
This in turn defines $N-1$ edges $(\mathbf{e}^i)_{i=0}^{N-2}$, with each edge given by $\mathbf{e}^i = \mathbf{q}_{i+1} - \mathbf{q}_i$.
For clarity, subscript indices refer to vertex-related quantities, while superscript indices refer to edge-related quantities.

To capture bending and twisting effects, two orthonormal frames are defined for each edge: a reference frame $\{\mathbf{d}^i_1, \mathbf{d}^i_2, \mathbf{t}^i\}$ and a material frame $\{\mathbf{m}^i_1, \mathbf{m}^i_2, \mathbf{t}^i\}$, as illustrated in Fig.~\ref{fig:centerline}.
The shared tangent director is given by $\mathbf{t}^i = \mathbf{e}^i / \lVert \mathbf{e}^i \rVert$.
The material frame is used to compute the discrete curvatures along the centerline and is obtained by rotating the reference frame about $\mathbf{t}^i$ by a signed angle $\theta^i$.
Without reference frames, each material frame would have to be computed directly from the preceding edge’s material frame, resulting in a dense Jacobian.
By introducing reference frames, the resulting elastic force Jacobian remains banded, which makes the implicit solve significantly more efficient.
Details on the temporal update of the reference frame can be found in~\cite{bergou2010dvt}.
Overall, $N$ vertices and $N-1$ edges correspond to a total of $4N-1$ degrees of freedom (DOF):
$\mathbf{x} = [\mathbf{q}_0, \theta^0, \mathbf{q}_1, \ldots, \theta^{N-2}, \mathbf{q}_{N-1}]^T$.

\subsection{Elastic Energies}
\textsc{DisMech} follows the Kirchhoff-rod assumption used in DER-based formulations, in which transverse shear deformation is neglected~\cite{bergou2008der, bergou2010dvt}. 
Shear effects may become non-negligible for short or thick segments of soft bodies with low effective shear stiffness. In such cases, the present framework can be extended by replacing the Kirchhoff strain description with a shearable Cosserat-rod or variable-strain formulation, as used in \textsc{Elastica}~\cite{gazzola2018forward} and SoroSim~\cite{mathew2022sorosim}.

We next briefly formulate the stretching, bending, and twisting energy terms, denoted by $E_s$, $E_b$, and $E_t$.

\bl{Stretching Energy.}
Stretching deformations are computed from the extension or compression of each edge.
Let $\bar{\lambda}^i$ denote the rest length of edge $i$.
Hereafter, an overbar denotes a quantity in the rest configuration.
The stretching strain of edge $i$ is defined as
\begin{equation}
    \epsilon^i = \frac{\lVert \mathbf e^i \rVert}{\bar{\lambda}^i} - 1.
\end{equation}
The total stretching energy is then
\begin{equation}
    E_s = \dfrac{1}{2} K_s \sum_{i=0}^{N-2} \left(\epsilon^i\right)^2 \bar{\lambda}^i,
\end{equation}
where $K_s$ is the stretching stiffness.

\bl{Bending Energy.}
Unlike stretching, bending deformation occurs between adjacent edges and can be quantified by the misalignment of their tangents.
At each interior node, we define the curvature binormal as
\begin{equation}
    \left(\kappa\mathbf b\right)_i = \dfrac{2 \mathbf t^{i-1} \times \mathbf t^i}{1 + \mathbf t^{i-1} \cdot \mathbf t^i}.
\end{equation}
Using this quantity, the integrated curvature components along the material frame basis vectors are
\begin{align*}
   \kappa_{1,i} &= \frac{1}{2}(\kappa \mathbf b)_i \cdot \left( \mathbf m_2^{i-1} + \mathbf m_2^i \right),  \\
   \kappa_{2,i} &= -\frac{1}{2}(\kappa \mathbf b)_i \cdot \left( \mathbf m_1^{i-1} + \mathbf m_1^i \right),
\end{align*}
resulting in $\boldsymbol{\kappa}_i = [\kappa_{1,i}, \kappa_{2,i}]^T$.
The bending energy is then given by
\begin{equation}
    E_b = \dfrac{1}{2} K_b \sum^{N-2}_{i=1} \left\| \boldsymbol{\kappa}_i - \bar{\boldsymbol{\kappa}}_i \right\|^2 \dfrac{1}{V_i},
\end{equation}
where $K_b$ is the bending stiffness and $V_i$ is the Voronoi length of the $i$-th interior node.

\bl{Twisting Energy.}
The final deformation mode is twisting, which measures the relative rotation between adjacent material frames.
As with bending, twisting is evaluated only at interior nodes using the material and reference frames of neighboring edges.
The integrated twist at interior node $i$ is defined as
\begin{equation}
    \psi_i = \theta^i - \theta^{i-1} + \beta^i,
\end{equation}
where $\beta^i$ denotes the signed angle offset between the reference frames of consecutive edges.
The twisting energy is then
\begin{equation}
    E_t = \dfrac{1}{2} K_t \sum^{N-2}_{i=1} \left( \psi_i - \bar{\psi}_i \right)^2 \dfrac{1}{V_i},
\end{equation}
where $K_t$ is the twisting stiffness.

Combining all three deformation modes, the total elastic energy is
\begin{equation}
    E_{\mathrm{elastic}} = E_s + E_b + E_t.
\end{equation}

\subsection{Delta Natural Curvature Control}

As first introduced in~\cite{choi_dismech}, we explicitly control the spatial bending deformation of the elastic rod by manipulating the natural curvature $\bar{\boldsymbol{\kappa}}_i$ at selected points along the rod.
For a naturally straight rod, $\bar{\boldsymbol{\kappa}}_i = [0, 0]^T$.
Incremental control is achieved by updating $\bar{\boldsymbol{\kappa}}_i$ through a control input $\Delta \bar{\boldsymbol{\kappa}}_i$.
With sufficiently high bending stiffness $K_b$, this yields highly responsive actuation, as shown in Fig.~\ref{fig:snapshot}.
If the rod has $C$ control points, this corresponds to $C$ DOFs in the 2D case and $2C$ DOFs in the 3D case.
We can further expand the soft manipulator’s workspace through natural twist control, where twisting deformation is generated by modulating the natural twist $\bar{\psi}_i$.
For full 3D actuation combining bending and twist, this results in a total of $3C$ DOFs.

In the presence of gravity, natural curvature and twist control play a role similar to PD joint position control in rigid manipulators.
Intuitively, 2D bending behaves like a revolute joint about an axis orthogonal to the link, 3D bending like a spherical joint, and twisting like a revolute joint collinear with the link.
This yields a smooth, low-dimensional, and physically interpretable action space for RL.

\section{Experiments}
\label{sec:experiments}

We now present results for four soft manipulation control tasks originally introduced in~\cite{Naughton2021_elastica_rl}.
Below we list several pragmatic adjustments made to better aid sim-to-real transfer:
\begin{enumerate}[leftmargin=12pt, topsep=0pt, itemsep=0pt]
    \item As previously discussed, \textbf{delta natural curvature and delta twist control} are now used in place of the original torque control approach in~\cite{Naughton2021_elastica_rl}. This choice makes the learned policies more amenable to successful sim-to-real transfer, as torque control typically exhibits a higher sim-to-real gap than delta position control~\cite{aljabout2024actionspace}.
    \item \textbf{Gravity} is included in all tasks, providing both a more realistic simulation environment and introducing additional nonlinearity to the control problem.
    \item The \textbf{control frequency} is set to 2\,Hz for contact tasks and 10\,Hz for non-contact tasks, in contrast to the original 3500\,Hz frequency used in~\cite{Naughton2021_elastica_rl}.
    Such high control frequencies often exacerbate the sim-to-real gap due to compounding errors. 
    We hypothesize that this high control frequency was originally chosen because of the large computational cost of time-stepping in \textsc{Elastica}. 
    Due to this cost, the interval between RL actions must be kept short; otherwise, training times become prohibitively long (as discussed later in Fig.~\ref{fig:results}).
\end{enumerate}

\bl{Training Parameters.} Given the relatively high computational cost of soft body simulation, we opt for an off-policy algorithm to maximize sample efficiency and choose the tried and true Soft Actor-Critic (SAC) algorithm~\cite{haarnoja_sac_2018}. 
Across all tasks, we use the same learning parameters listed in Table~\ref{tab:train_params} to demonstrate robustness to hyperparameter tuning.

\bl{Simulation Parameters.} For all tasks, \textsc{DisMech} uses a simulation timestep size of $\Delta t=0.05$\,s while \textsc{Elastica} uses $\Delta t =0.0002$\,s.
For non-contact tasks, a control frequency of 10\,Hz is used, resulting in \textsc{DisMech} and \textsc{Elastica} performing one control step every 2 and 500 sim steps, respectively.
For contact tasks, the control frequency is reduced to 2\,Hz to coarsen the action space and aid exploration.
This corresponds to one control step every 10 and 2500 simulation steps for \textsc{DisMech} and \textsc{Elastica}, respectively.
A summary of all sim-related parameters are listed in Table~\ref{tab:sim_params}.

\bl{Control Setup.} Across the soft manipulator, five approximately equidistant control points are used.
With five control points, this then results in a total of 5, 10, and 15 DOF for 2D bending, 3D bending, and 3D bending + twist, respectively.
To prevent sharp bending or twisting, the delta control signals are spatially smoothed within each control point's Voronoi region.
To ensure smooth actuation, we temporally smooth delta commands by linearly interpolating the control signal across the simulation steps.
The same geometric and material parameters for the soft manipulator are used across all tasks, as listed in Table~\ref{tab:soft_manipulator_parameters}.

\subsection{Results and Wall-clock Time Comparison}

\begin{figure*}[!t]
    \centering
	\includegraphics[width=0.95\textwidth]{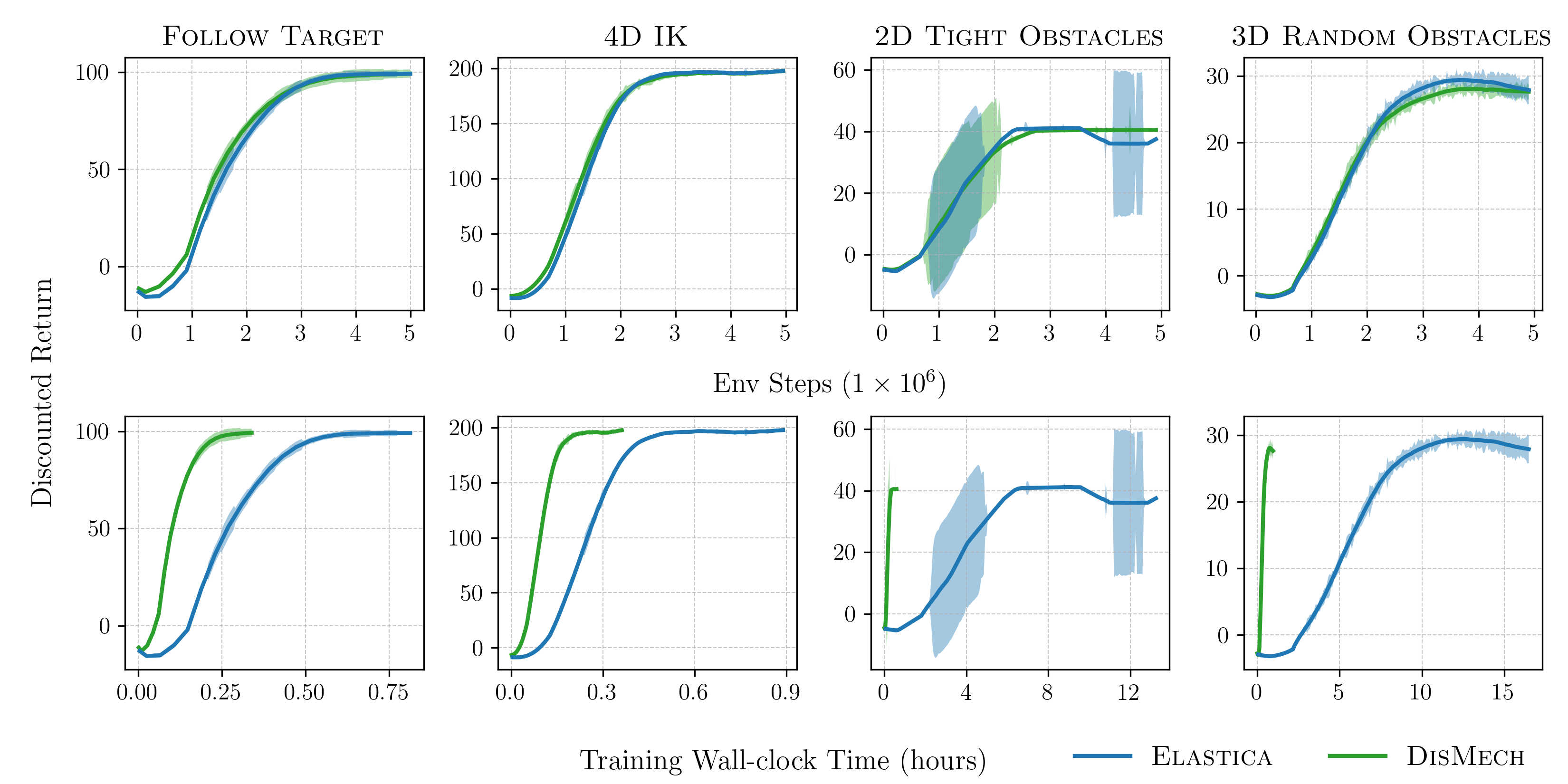}
    	\caption{Comparison of the average discounted return (Eq.~\ref{eq:avg_discounted_return}) with respect to bo
        th environment steps (top row) and training wall-clock time (bottom row) across five random seeds. For each task, we can see almost identical convergence rates regardless of the choice of simulator with respect to environment steps. Minimal variance can be observed aside from \textsc{2D Tight Obstacles}, given the narrow success condition. Despite the similar convergence properties, plotting the same data against wall-clock time shows the immense speed benefits of leveraging implicit time-stepping.} 
\label{fig:results}
\end{figure*}

Training experiments are conducted for each task using both \textsc{Elastica} and \textsc{DisMech}. For each simulator and task, we report results across five random seeds using 500 parallel environments.
All experiments are performed on a 128-core Intel Xeon Platinum 8369B @ 2.90\,GHz CPU and an NVIDIA GeForce RTX 3090 GPU. 
The average discounted return (Eq.~\ref{eq:avg_discounted_return}) with respect to both environment steps and wall-clock time is shown in Fig.~\ref{fig:results}, and the corresponding average wall-clock times are summarized in Table~\ref{tab:time}.

\bl{Task 1:} \textsc{Follow Target.} In this non-contact task, the soft manipulator is required to follow a moving target point with its end effector using 3D bending control, as shown in the top row of Fig.~\ref{fig:snapshot}.
Between the two simulators, we observe similar convergence behavior in Fig.~\ref{fig:results} with respect to the number of environment steps.
In terms of total training wall-clock time, \textsc{DisMech} achieves a $2.4\times$ improvement over \textsc{Elastica}, primarily due to a $6.21\times$ speedup in parallel stepping.

\bl{Task 2:} \textsc{4D IK.} In this non-contact task, a random 4D end-effector target pose $(x, y, z, \theta_\textrm{yaw})$ target is sampled, forming an inverse kinematics problem for the soft manipulator.
Given the yaw target $\theta_\textrm{yaw}$, delta natural twist control is used along with 3D bending.
Once again, we observe nearly identical convergence with respect to the number of environment steps, with results consistent with those of \textsc{Follow Target}: a $2.45\times$ improvement in training wall-clock time driven by a $5.69\times$ speedup in parallel stepping.

\bl{Task 3:} \textsc{2D Tight Obstacles.} The first of our contact tasks, Task~3 requires the soft manipulator to reach a static target position located beyond rigid obstacles in 2D.
Both the target position and the obstacle configuration remain fixed across episodes, but the available clearance is extremely tight (0.12\,m clearance for a 0.1\,m diameter manipulator), posing a challenging exploration problem.
The control type used is 2D bending.

Contact is notoriously difficult to simulate due to its stiff nonlinear nature, and with its inclusion we begin to see the advantages of implicit time-stepping magnified severalfold.
\textsc{Elastica} handles contact by computing a repulsive force
\begin{equation}
\mathbf{F}_\textrm{contact}^\textrm{elastica} = H(\epsilon)(-\mathbf{F}_\perp + k\epsilon + \mathbf{d})\hat{\mathbf{u}},
\label{eq:elastica_contact}
\end{equation}
where $\epsilon$ is the penetration distance (negative when penetrating), $H(\cdot)$ is the Heaviside function, $k$ is the contact stiffness, $\mathbf{d}$ is a damping force, and $\hat{\mathbf{u}}$ is the contact normal.
Force equilibrium is maintained as $-\mathbf{F}_\perp$ represents the sum of all forces pushing the rod against the contact surface.

In contrast, \textsc{DisMech} employs a fully implicit contact formulation—the Implicit Contact Method (IMC)~\cite{choi_imc_2021, tong_imc_2022}—which allows for large, aggressive timestep sizes despite the stiffness of contact dynamics.
With this formulation, we observe an overall training speedup of $22.45\times$, driven by a $39.78\times$ improvement in parallel stepping.

\bl{Task 4:} \textsc{3D Random Obstacles.} In this final task, the soft manipulator must reach a static target position located beyond eight rigid obstacles in 3D, as shown in the middle row of Fig.~\ref{fig:snapshot}.
Although the target position remains fixed across episodes, the rigid obstacles are randomized.
Their positions and orientations are sampled within a 3D bounding box in a contact-free manner.
Given this sampling scheme, reaching the target is not always feasible.
The control type used is 3D bending.

Similar to \textsc{2D Tight Obstacles}, we again observe a substantial wall-clock speedup of $17.24\times$ in overall training, resulting from a $24.16\times$ improvement in parallel stepping.

\begin{table*}[!t]
\renewcommand{\arraystretch}{1.1}
\caption{Wall-clock comparison using 500 parallel environments. The left column reports the average wall-clock time to complete one full training iteration, while the right column reports the average wall-clock time to complete one full parallel environment step. As all training parameters are identical except for the simulator itself, we can conclude that the significant overall training speedup is primarily driven by the faster parallel stepping performance.}
\centering
\small
\begin{tabular*}{\textwidth}{@{\extracolsep{\fill}} l cccc cccc}
\toprule
\multirow{2}{*}[-3pt]{Simulator} & \multicolumn{4}{c}{Train Iter Time (ms) [$\downarrow$]} & \multicolumn{4}{c}{Parallel Env Step Time (ms) [$\downarrow$]} \\
\cmidrule(lr){2-5}
\cmidrule(lr){6-9}
& \textsc{Fol Tar} & \textsc{4D IK} & \textsc{2D Obs} & \textsc{3D Obs} & \textsc{Fol Tar} & \textsc{4D IK} & \textsc{2D Obs} & \textsc{3D Obs}\\
\midrule

\textsc{Elastica} & $292.69$ & $320.27$ & $4772.27$ & $5930.42$ & $191.66$ & $215.89$ & $4671.93$ & $5816.21$ \\
\textsc{DisMech} & $122.11$ & $130.48$ & $212.59$ & $344.04$ & $30.88$ & $37.22$ & $117.44$ & $240.70$ \\
\cmidrule{1-9}
Speedup & $2.40\times$ & $2.45\times$ & $22.45\times$ & $17.24\times$ & $6.21\times$ & $5.69\times$ & $39.78\times$ & $24.16\times$ \\
\bottomrule
\end{tabular*}
\label{tab:time}
\end{table*}

\subsection{Sim-to-sim Analysis}

\begin{table*}[!t]
\renewcommand{\arraystretch}{1.1}
\caption{Sim-to-sim comparison of the average discounted return for policies across different evaluation simulators. Evaluation is done by averaging results across 100 episodes for each of the five random seed policies, resulting in 500 episodes total. $^*$ indicates that \textsc{DisMech}'s contact force had to be tuned for the \textsc{Elastica} trained policy to be able to reach optimal performance. Note that \textsc{DisMech} trained policies were able to reach optimal performance without any tuning.}
\centering
\small
\begin{tabular*}{\textwidth}{@{\extracolsep{\fill}} l cccc}
\toprule
\multicolumn{5}{c}{Average Discounted Return [$\uparrow$]} \\
\cmidrule{1-5}
Train Sim $\rightarrow$ Eval Sim & \textsc{Fol Tar} & \textsc{4D IK} & \textsc{2D Obs} & \textsc{3D Obs} \\
\midrule
\textsc{Elastica} $\rightarrow$ \textsc{Elastica} & \meanstd{95.14}{8.40} & \meanstd{183.62}{63.66} & \meanstd{41.23}{2.97} & \meanstd{27.50}{19.24} \\
\textsc{Elastica} $\rightarrow$ \textsc{DisMech} & \meanstd{87.08}{7.98} & \meanstd{179.32}{65.15} & \meanstd{40.33}{5.45}$^*$ & \meanstd{27.31}{18.42}$^*$ \\
\textsc{DisMech} $\rightarrow$ \textsc{DisMech} & \meanstd{96.26}{9.37} & \meanstd{186.24}{63.18} & \meanstd{40.91}{2.98} & \meanstd{27.17}{19.03} \\
\textsc{DisMech} $\rightarrow$ \textsc{Elastica} & \meanstd{90.26}{9.37} & \meanstd{185.96}{62.39} & \meanstd{40.69}{2.99} & \meanstd{28.05}{18.04} \\
\bottomrule
\end{tabular*}
\label{tab:sim2sim}
\end{table*}

In addition to comparing training convergence and wall-clock speed between the simulators, we also conduct an extensive sim-to-sim evaluation to examine any possible dynamics differences between them.
To this end, we evaluate each of the five \textsc{Elastica} and \textsc{DisMech} policies for each task within both simulators.
The corresponding discounted returns are reported in Table~\ref{tab:sim2sim}.

Across all tasks, the mean and standard deviation of the discounted return differ only marginally, indicating close agreement between the two simulators.
One major exception, however, arises in the \textsc{Elastica} $\rightarrow$ \textsc{DisMech} evaluations for the contact-rich tasks.
For these evaluations, we observed a significant drop in discounted return with \textsc{2D Tight Obstacles} failing completely (negative return) and \textsc{3D Random Obstacles} exhibiting an over 40$\%$ decrease in performance.

We attribute this large decrease in performance to the fact that \textsc{Elastica}’s contact formulation (Eq.~\ref{eq:elastica_contact}) often produced spongy contact behavior, where resistance began slightly above the surface and noticeable penetrations occurred even when using the same contact stiffness of $k = 1.6\times10^5$ as in~\cite{Naughton2021_elastica_rl}.
This behavior is expected given the simple linear spring term $k\epsilon$ in Eq.~\ref{eq:elastica_contact}.
In contrast, \textsc{DisMech} employs a fully implicit contact formulation, the Implicit Contact Method (IMC)~\cite{choi_imc_2021, tong_imc_2022}, whose force is derived from a squared potential:
\begin{equation}
\mathbf{F}_\mathrm{contact}^\mathrm{dismech} = -k \nabla_\mathbf{q}
\left(
\dfrac{1}{K} \log \left( 1 + e^{K\epsilon} \right)
\right)^{2},
\end{equation}
where $K = 15/\delta$ and $\delta$ is a contact distance tolerance.
Although this is also a form of soft contact, the squared term produces quadratically higher resistance as a function of $\epsilon$, while $\delta$ directly controls the stiffness of the contact energy curve.
By tuning $\delta$ until the contact response was visually rigid, \textsc{DisMech} achieved noticeably more realistic contact compared to \textsc{Elastica} (Fig.~\ref{fig:contact_comparison}).
In the end, the similar return values labeled with $^*$ in Table~\ref{tab:sim2sim} were obtained only after relaxing the \textsc{DisMech} contact parameters $k$ and $\delta$ to approximate the softer contact dynamics of \textsc{Elastica}.
In contrast, \textsc{DisMech} $\rightarrow$ \textsc{Elastica} evaluations maintained comparable performance despite the latter’s softer contact, suggesting that \textsc{DisMech}’s tighter contact bounds make control tasks more challenging rather than easier.

Finally, \textsc{Elastica} includes shear deformation in its Cosserat-rod formulation, whereas the current \textsc{DisMech} implementation uses a Kirchhoff/DER model. The close sim-to-sim agreement in Table II suggests that shear is not a dominant effect for the benchmarked slender-manipulator tasks. We therefore attribute the observed discrepancies primarily to contact modeling rather than to the underlying elastic rod representation.

\section{Conclusion and Future Work}
\label{sec:conclusion}

In this work, we conducted an extensive comparison of soft robot policy learning using the fully implicit simulator \textsc{DisMech} and the explicit simulator \textsc{Elastica} across four representative manipulation tasks.
We also introduced \textit{delta natural curvature control} as an intuitive and efficient method for controlling soft, rod-like robots. 
Despite both simulators relying on similar reduced-order centerline formulations, \textsc{DisMech} consistently achieved comparable control performance while offering up to $22\times$ faster training and nearly $40\times$ faster parallel stepping.
Sim-to-sim evaluations further confirmed that \textsc{DisMech} closely reproduces \textsc{Elastica}’s dynamics, with differences arising primarily from contact modeling rather than elastic behavior.

Several exciting avenues remain for future work.
The most immediate direction is to transition \textsc{DisMech} from CPU to GPU. 
GPU implementation will be key to achieving the large-scale parallelization seen in rigid-body simulators and would make on-policy learning substantially more feasible.
Finally, effective sim-to-real transfer strategies for delta natural curvature policies must be developed and validated through real robot deployment.
To accomplish this, a sim-to-real mapping function $f(\Delta \bar{\boldsymbol \kappa}_i)$ that translates delta natural curvature actions into real, hardware-specific inputs (e.g., voltage for shape memory alloy (SMA) actuators), must be established.
For quasi-static behaviors, a simple data-driven mapping from delta curvature commands to actuator signals may suffice, whereas dynamic control will require actuator-aware modeling of SMA dynamics.
Conditioning on the history may also be necessary in order to model the hysteresis effect common in real soft manipulators.

\bibliography{references}
\bibliographystyle{IEEEtran}

\clearpage
\appendix
\renewcommand{\thetable}{A.\arabic{table}}
\setcounter{table}{0}
\renewcommand{\thefigure}{A.\arabic{figure}}
\setcounter{figure}{0}

\bl{Markov Decision Process.}
We formulate each task as a Markov Decision Process (MDP)
$(\mathcal{S}, \mathcal{A}, P, r, \gamma)$, where $\mathcal{S}$ is the state
space, $\mathcal{A}$ is the action space, $P(s_{t+1}\mid s_t,a_t)$ denotes the
transition dynamics, $r_{t+1} = r(s_t,a_t)$ is the reward received after
taking action $a_t$, and $\gamma \in [0,1]$ is the discount factor.
A policy $\pi(a_t \mid s_t)$ induces a trajectory
$(s_0,a_0,r_1,s_1,\ldots)$ over an episode consisting of $T$ transitions.

The discounted return from time step $t=0$ is defined as
\begin{equation}
    R = \sum_{t=0}^{T-1} \gamma^t r_{t+1}.
\end{equation}

\bl{Average Discounted Return.} In this work, we report the \emph{average discounted return}, which normalizes
the discounted return by the episode length. For an episode consisting of $T$
transitions, it is defined as
\begin{equation}
    \bar{R} = \frac{1}{T} R.
\end{equation}

Each training run uses $W$ parallel environments. For a given random seed,
we compute the mean average discounted return across all parallel environments:
\begin{equation}
    \bar{R}_{\text{seed}} = \frac{1}{W} \sum_{w=0}^{W-1} \bar{R}_w,
\label{eq:avg_discounted_return}
\end{equation}
where $\bar{R}_w$ denotes the average discounted return of environment $w$.
All results in the main paper report the mean and standard deviation of
$\bar{R}_{\text{seed}}$ aggregated across random seeds.
In this work, we directly reused the reward formulations from~\cite{Naughton2021_elastica_rl}.
These can be found listed in detail at~{\tt\url{https://www.cosseratrods.org/Elastica+RL/}}.

\bl{Training Parameters.} The following training parameters are used across all tasks.
For SAC, we opt to exclude entropy from the reward formulation as doing so can often lead to suboptimal learning~\cite{yu2022needentropyrewardin}.
Both actor and critic networks are simple MLPs and use the same hidden dimension size.

\begin{table}[h]
\renewcommand{\arraystretch}{1.1}
\caption{Training Parameters}
\centering
\footnotesize
\begin{tabular}{ll}
\toprule
Parameter & Value \\
\midrule
Number of envs $W$ & 500 \\
Update-to-data (UTD) ratio & 4 \\
Hidden dim size & (256, 256, 256) \\
SAC soft update coefficient $\tau$ & 0.005 \\
SAC soft update period & 8 \\
Discount factor $\gamma$ & 0.99 \\
Entropy reward & None \\
Optimizer & Adam \\
Learning rate & 0.001 \\
Mini batch size & 2048 \\
Replay buffer length & 2,000,000 \\
\bottomrule
\end{tabular}
\label{tab:train_params}
\end{table}

\bl{Simulation Parameters.} The following simulation parameters are used across all tasks. Parameters are split into common and simulator-specific sections.

\begin{table}[H]
\renewcommand{\arraystretch}{1.1}
\caption{Simulation Parameters}
\centering
\footnotesize
\begin{tabular}{ll}
\toprule
Common Parameter & Value \\
\midrule
Number of discrete nodes $N$ & 21 \\
Number of control points $C$ & 5 \\
\midrule
\textsc{Elastica} Parameter & Value \\
\midrule
Time-step $\Delta t$ & 0.0002\,s \\
Control period (non-contact) & 500 \\
Control period (contact) & 2500 \\
Contact stiffness $k$ & 1.6e5 \\
\midrule
\textsc{DisMech} Parameter & Value \\
\midrule
Time-step $\Delta t$ & 0.05\,s \\
Control period (non-contact) & 2 \\
Control period (contact)  & 10 \\
Contact stiffness $k$ & 1e6 \\
Contact distance tolerance $\delta$ & 0.005\,m \\
Max Newton iterations (non-contact) & 2 \\
Max Newton iterations (contact) & 5 \\
\bottomrule
\end{tabular}
\label{tab:sim_params}
\end{table}

\bl{Soft Manipulator Parameters.} The following geometric and material parameters are used across all tasks.

\begin{table}[H]
\renewcommand{\arraystretch}{1.1}
\caption{Soft Manipulator Parameters}
\centering
\footnotesize
\begin{tabular}{ll}
\toprule
Parameter & Value \\
\midrule
Length $L$ & 1.0\,m \\
Radius $r$ & 0.05\,m \\
Density $\rho$ & 1000\,kg/m$^3$ \\
Young's modulus $E$ & 10\,MPa \\
Poisson's ratio $\nu$ & 0.5 \\
\bottomrule
\end{tabular}
\label{tab:soft_manipulator_parameters}
\end{table}

\begin{figure}[h]
\centering
\includegraphics[width=\columnwidth]{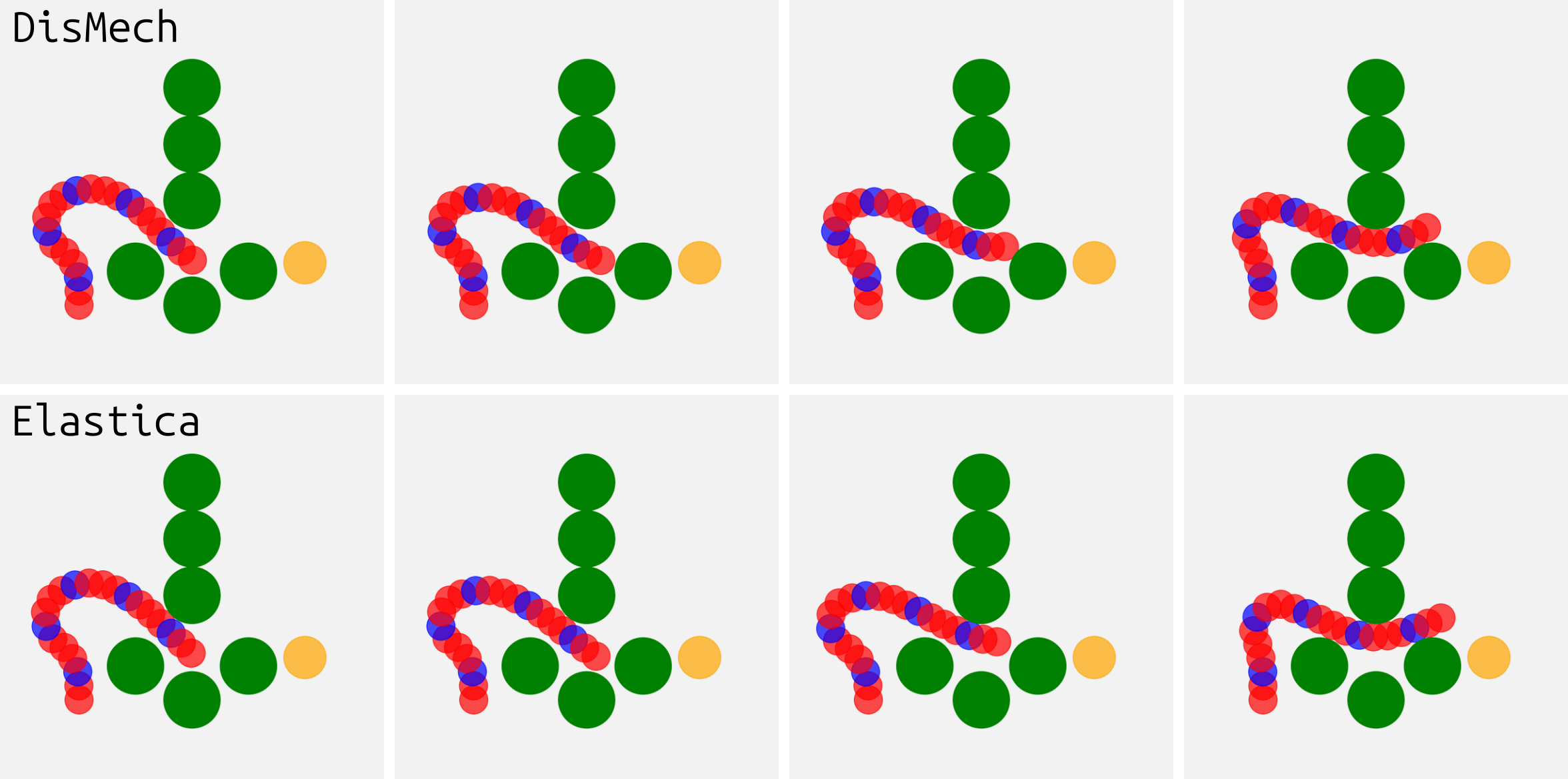}
\caption{Qualitative comparison of contact behavior between \textsc{DisMech} and \textsc{Elastica} for the \textsc{2D Tight Obstacles} task. The same policy is evaluated in both simulators. Note the noticeable gaps between the end-effector and the obstacles during insertion into the tight crevice that are present in \textsc{Elastica}, but not in \textsc{DisMech}, due to the soft spring contact model in Eq.~\ref{eq:elastica_contact}. In contrast, \textsc{DisMech}'s contact formulation enforces non-penetration directly at the contact surface, resulting in continuous contact without visible gaps.}
\label{fig:contact_comparison}
\end{figure}

\end{document}